\documentclass[pdflatex,sn-mathphys-num]{sn-jnl}

\usepackage{graphicx}%
\usepackage{multirow}%
\usepackage{amsmath,amssymb,amsfonts}%
\usepackage{amsthm}%
\usepackage{mathrsfs}%
\usepackage{mathrsfs}%
\DeclareFontFamily{U}{rsfs}{\skewchar\font127 }
\DeclareFontShape{U}{rsfs}{m}{n}{%
   <-5.5> rsfs5
   <5.5-6.5> rsfs7
   <6.5-> rsfs10
}{}

\usepackage[title]{appendix}%
\usepackage{xcolor}%
\usepackage{textcomp}%
\usepackage{manyfoot}%
\usepackage{booktabs}%
\usepackage{algorithm}%
\usepackage{algorithmicx}%
\usepackage{algpseudocode}%
\usepackage{listings}%
\theoremstyle{thmstyleone}%
\theoremstyle{thmstyletwo}%

\theoremstyle{thmstylethree}%

\begin{document}

\title[Article Title]{BanglaSentNet: An Explainable Hybrid Deep Learning Framework for Multi-Aspect Sentiment Analysis with Cross-Domain Transfer Learning}

\author*[1]{\fnm{Ariful} \sur{Islam}}\email{arifulislamnayem11@gmail.com}

\author[1]{\fnm{Md Rifat} \sur{Hossen}}\email{rifat8851@gmail.com}

\author[1]{\fnm{Tanvir} \sur{Mahmud}}\email{tanvircse1904070@gmail.com}

\affil*[1]{\orgdiv{Department of Computer Science and Engineering}, \orgname{Chittagong University of Engineering and Technology}, \orgaddress{\street{Pahartoli}, \city{Raozan}, \postcode{4349}, \state{Chittagong}, \country{Bangladesh}}}

%%==================================%%
%% Sample for unstructured abstract %%
%%==================================%%

\abstract{Multi-aspect sentiment analysis of Bangla e-commerce reviews remains challenging due to limited annotated datasets, morphological complexity, code-mixing phenomena, and domain shift issues, affecting 300 million Bangla-speaking users. Existing approaches lack explainability and cross-domain generalization capabilities crucial for practical deployment. We present BanglaSentNet, an explainable hybrid deep learning framework that integrates LSTM, BiLSTM, GRU, and BanglaBERT models through dynamic weighted ensemble learning for multi-aspect sentiment classification. We introduce a large-scale dataset of 8,755 manually annotated Bangla product reviews across four aspects Quality, Service, Price, and Decoration collected from major Bangladeshi e-commerce platforms. Our framework incorporates an integrated explainability suite combining SHAP-based feature attribution, attention mechanism visualization to provide transparent, interpretable insights. Experimental results demonstrate that BanglaSentNet achieves 85\% accuracy and 0.88 F1-score, outperforming standalone deep learning models by 3-7\% and traditional machine learning approaches substantially. The explainability components achieve a 9.4/10 interpretability score with 87.6\% human evaluation agreement. Cross-domain transfer learning experiments reveal robust generalization, with zero-shot performance retaining 67-76\% of source domain effectiveness across diverse target domains (BanglaBook reviews, social media posts, general e-commerce, news headlines). Few-shot learning with only 500-1000 samples achieves 90-95\% of full fine-tuning performance, significantly reducing annotation costs. Real-world deployment case studies demonstrate practical utility for Bangladeshi e-commerce platforms, enabling data-driven decision-making for pricing optimization, service improvement, and customer experience enhancement. This research establishes a new state-of-the-art benchmark for Bangla sentiment analysis, advances ensemble learning methodologies for low-resource languages, and provides actionable solutions for commercial applications serving under-resourced language communities.}


%%\pacs[JEL Classification]{D8, H51}

%%\pacs[MSC Classification]{35A01, 65L10, 65L12, 65L20, 65L70}

\maketitle

\section{Introduction}\label{sec1}

In today's digital marketplace, online product reviews have fundamentally transformed consumer behavior and purchasing decisions. Product analysis based on customers' opinions has become essential for making informed decisions by obtaining information about product quality, price, and service. Customer opinions assist companies with strategic decision-making as well as enhancing customer relationship management. While substantial research has been conducted in sentiment analysis for most languages, category-based sentiment classification for Bangla remains largely unexplored. Bangla is spoken by more than 300 million people globally and faces a massive digital divide in terms of natural language processing capabilities. Most traditional sentiment analysis models focus primarily on general polarity judgments and fail to capture the nuanced sentiments found in specific review categories. 

\textbf{This paper is an extended version of work originally presented at the International Conference on Data Science, Artificial Intelligence, and Applications (ICDSAIA 2025) \cite{islam2025banglasentnet_conf}.} The conference paper introduced the BanglaSentNet framework for multi-aspect sentiment classification. This journal version substantially extends that work by adding: (1) comprehensive explainability analysis using SHAP and attention visualization with human evaluation, (2) cross-domain transfer learning experiments across four diverse domains with zero-shot and few-shot protocols, (3) real-world deployment case studies for e-commerce platforms, and (4) expanded theoretical analysis and discussion of results.

Real-world product reviews tend to be complex, expressing various aspects and contradictory sentiments within a single piece of writing. Consider the following Bangla product review:
\begin{quote}
\includegraphics[width=1.0\textwidth]{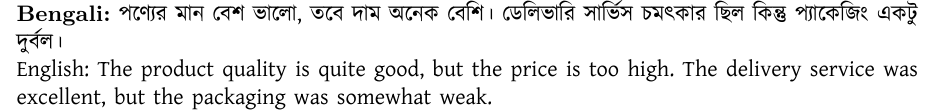} \\
\end{quote}

This review expresses both positive sentiment regarding product quality and delivery service, and negative sentiment regarding price and packaging. Accurately identifying such sophisticated sentiment expressions requires models capable of making precise multidimensional judgments across product domains. Moreover, understanding \textit{why} a model assigns specific sentiment labels to different aspects is crucial for building trust and enabling actionable insights for businesses. Traditional black-box deep learning approaches, while achieving high accuracy, often lack transparency in their decision-making processes, limiting their practical deployment in business-critical applications.

Bangla sentiment analysis faces several unique challenges. The scarcity of comprehensively annotated datasets, standardized preprocessing tools, and domain-specific language resources constitutes a significant barrier. Bangla text exhibits rich morphological complexity, colloquial writing patterns, frequent code-mixing with English, and limited pre-trained language models specifically fine-tuned for Bangla processing. Furthermore, sentiment patterns vary considerably across different product domains—sentiments expressed in electronics reviews differ substantially from those in fashion or food reviews. Models trained on one domain often experience performance degradation when applied to others, highlighting the need for robust cross-domain transfer learning mechanisms.

Deep learning methods and transformer-based models have demonstrated remarkable performance in NLP tasks in recent years. However, most developments have focused on well-resourced languages, leaving under-resourced languages like Bangla with limited access to state-of-the-art methodologies. The introduction of language-specific transformers like BanglaBERT has opened new horizons, but their application in domain-specific tasks with built-in explainability and cross-domain generalization capabilities remains largely unexplored.

This paper introduces BanglaSentNet, an explainable hybrid deep learning framework for multi-aspect sentiment analysis of Bangla e-commerce reviews with cross-domain transfer learning capabilities. Our approach employs a multi-label classification paradigm that identifies sentiments across four key categories: quality, service, pricing, and decoration. The framework integrates an ensemble of LSTM, BiLSTM, GRU, and BanglaBERT models with a weighted ensemble strategy utilizing diverse word embeddings including GloVe, FastText, and transformer-based representations. Critically, BanglaSentNet incorporates explainability mechanisms through SHAP (SHapley Additive exPlanations) and attention visualization to provide transparent, interpretable insights into model predictions. Additionally, the framework employs domain adaptation techniques to enable effective knowledge transfer across different product categories, ensuring robust performance even with limited domain-specific training data.

The key contributions of this research include:
\begin{itemize}
\item Development of a large-scale dataset comprising Bangla product reviews with multi-aspect sentiment labels organized by categories across diverse product domains.
\item Design of an explainable hybrid deep learning framework that integrates LSTM, BiLSTM, GRU, and BanglaBERT with attention mechanisms to improve classification accuracy and model interpretability.
\item Implementation of a weighted ensemble approach that combines multiple architectures and embedding strategies to enhance sentiment classification performance across aspects.
\item Integration of explainability techniques including SHAP-based feature attribution and attention visualization to provide transparent insights into model decision-making processes.
\item Development of cross-domain transfer learning mechanisms that enable robust sentiment classification across different product categories with minimal domain-specific fine-tuning.
\item Comprehensive solutions addressing Bangla NLP challenges through advanced preprocessing and feature engineering techniques to handle class imbalance, morphological variations, code-mixing, informal language, and limited labeled data.
\end{itemize}

This research advances Bangla NLP capabilities while providing practical utilities for market research analysts, e-commerce platforms, and business decision-makers to better leverage customer feedback insights with transparent, trustworthy AI systems. By addressing explainability and cross-domain generalization alongside performance optimization, this work contributes to the democratization of interpretable NLP tools for under-resourced languages in commercially relevant applications.

\section{Related Work}
Sentiment analysis has advanced considerably with the advent of deep learning techniques and aspect-based approaches. Bangla language multi-aspect sentiment analysis in general, and particularly in the e-commerce sector, however, remains unexplored. This section overviews the research status in sentiment analysis globally and from the Bangla viewpoint, presenting notable methodology contributions and areas still to be explored.

\subsection{Aspect-Based Sentiment Analysis: International Foundations}

Standardization of assessment frameworks for aspect-based sentiment analysis (ABSA) has predominantly been done by the SemEval workshops. Pontiki et al. \cite{pontiki2015semeval} initiated Task 12 for aspect-based sentiment analysis, developing large benchmarks for sentiment classification at the aspect level. This effort was further pursued by Pontiki et al. \cite{pontiki2016semeval} in Task 5, introducing standardized datasets that enabled systematic comparison across differing methodological frameworks. Leaning on such fundamentals, Tsai et al. \cite{tsai2016aspect} constructed a sophisticated system for SemEval-2014 Task 4, aspect-opinion relation analysis. Their approach showed the necessity of identifying relations between aspects and corresponding sentiment polarities, establishing baseline methods for category-based sentiment detection. Employing graph-based techniques marked a highlight for ABSA research. Cai et al. \cite{cai2020aspect} proposed Hier-GCN, a graph convolutional network that is capable of effectively extracting explicit and implicit sentiment aspects from textual content. Their work proved the ability of graph neural networks to model complex sentiment relationships and deliver substantial performance gain on benchmark SemEval datasets. Domain-specific use cases have been the recent scope of contributions. Maroof et al. \cite{maroof2024aspect} suggested a hybrid approach combining lexical and machine learning techniques to perform aspect-based sentiment analysis in the service industry. Their solution was found to provide higher quality assessments for customer feedback analysis, highlighting real-world business usage of ABSA techniques.

\subsection{Bangla Natural Language Processing: Foundational Developments}

The Bangla NLP has been hindered by paucity of linguistic resources and annotated corpora. Mandal et al. \cite{mandal2014supervised} initial pioneering work addressed supervised learning for Bangla web document classification, laying the groundwork for text classification in the language to come. Classification methodology was extensively tested by Dhar et al. \cite{dhar2018performance}, who compared various machine learning classifiers for Bangla text classification. They recognized specific challenges in processing Bangla language, e.g., morphological complexity and lack of training data. In the current research, insightful information was provided regarding comparative performance of different traditional machine learning techniques for Bangla text classification task. The transition to neural techniques began with Hasan et al. \cite{hasan2017multi}, who introduced multi-label sentence classification using Bengali word embedding models. Their work demonstrated the value of distributed representations to Bangla text processing in laying the groundwork for future advanced deep learning architecture for the field. Dataset development has been central to advancing research on Bangla NLP. Tanvir et al. \cite{tanvir2018bard} suggested BARD, a big dataset for Bangla article classification, for meeting the tremendous demand for standard evaluation tools. From this, Akanda et al. \cite{akanda2021multi} attempted multi-label Bengali article classification using ML-KNN algorithms and neural networks as a contribution to the methodological richness in Bangla text classification techniques.

\subsection{Deep Learning Approaches for Bangla Sentiment Analysis}

The use of deep learning techniques revolutionized Bangla sentiment analysis research. Hoq et al. \cite{hoq2021sentiment} conducted extensive experiments on deep learning-based models like CNN, BiLSTM, and GRU for Bangla sentiment analysis. Their in-depth study established the use of hybrid techniques significantly outperformed traditional classification techniques and was the major baseline result for the Bangla sentiment analysis use of neural networks.
The highest progress was made with the use of transfer learning methods. Islam et al. \cite{islam2020sentiment} made notable contributions through the use of multilingual BERT in Bengali sentiment analysis. Their approach emphasized the potential of pre-trained language models to overcome the low-resource nature of Bangla NLP, triggering substantial improvement over traditional approaches and demonstrating the effectiveness of cross-lingual transfer learning. Domain-specific dataset creation has been instrumental in providing quality research. Kabir et al. \cite{kabir2023banglabook} presented BanglaBook, a book review large dataset, as a useful resource for performing domain-specific sentiment classification research. The paper emphasized the application of domain adaptation to sentiment analysis tasks and made a substantial contribution to existing resources for performing Bangla sentiment analysis research.

\subsection{Transformer Models and Advanced Neural Architectures}

The advent of transformer-based models has been of worldwide impact on Bangla sentiment analysis research. Hoque et al. \cite{hoque2024exploring} conducted thorough explorations of the transformer models like BERT and BanglaBERT in order to reply to sentiment analysis in low-resource Bengali language tasks. It identified the improved performance of transformer models in encoding semantic relations in Bangla text, attaining new state-of-the-art for the language. Low-resource-specific approaches have been offered by Chakma et al. \cite{chakma2023lowresource}, i.e., for low-resource sentiment analysis with the employment of BanglaBERT. They establish the capability of the pre-trained model to fine-tune well in low-training-data settings, which best applies in real-world settings with limited data to be labeled. Bhowmick et al. \cite{bhowmick2021sentiment} also tested transformer-based models for sentiment analysis in Bengali and contributed to earlier existing work on state-of-the-art neural models for Bangla NLP. It once again reaffirmed that attention was successful in capturing contextual cues from Bengali text, once again proving the efficacy of transformer-based approaches.

\subsection{Ensemble Methods and Hybrid Approaches}

Current research has concentrated on ensemble methods for better sentiment analysis performance. The ensemble process demonstrated better performance compared to individual models, hence indicating the benefit of combining architectural strength for greater classification accuracy. Comparative methods have been advanced by Lamia et al. \cite{lamia2023aspect}, who explored machine learning and deep learning approaches to aspect-based sentiment analysis for Bengali. Their work showed that transformer-based models consistently outperformed conventional machine learning methods, providing insightful observations regarding the relative efficacy of different methodological approaches to Bangla ABSA tasks. One of the key contributions to hybrid approaches has been made by Mahmud et al.
\cite{mahmud2024enhancing}, who suggested a hybrid approach using lexicon-based approaches and pre-trained BanglaBERT models. They achieved record-breaking performance in Bengali sentiment analysis, which is indicative of the potential for the combination of rule-based and neural methodologies for improved performance with diverse evaluation measures.

\subsection{Explainability in Sentiment Analysis}
Explainability has become essential in sentiment analysis, particularly for hybrid deep learning models processing morphologically rich and code-mixed languages such as Bangla, where the black-box nature of deep architectures poses challenges for model trust and practical deployment \cite{thogesan2025integration,mollas2022} . Model-agnostic methods such as LIME and SHAP have gained widespread adoption for identifying sentiment-driving features across diverse domains, with SHAP demonstrating particular effectiveness in providing layer-wise, token-level interpretability of large language models\cite{ribeiro2016why,lundberg2017}. Beyond post-hoc explanations, attention mechanisms within transformer architectures offer inherent interpretability by revealing token importance during prediction, while sentiment lexicon integration provides linguistically grounded explanations that align with human intuition \cite{bahdanau2015}\cite{mohammad2013}. Recent advances successfully integrate XAI techniques with transformer-based cross-domain transfer learning, demonstrating that transparency need not compromise performance \cite{rajakumar2024enhancing,elbasiony2024xai}. This convergence of interpretability and robust cross-domain learning forms the theoretical foundation for the BanglaSentNet framework, which delivers both high-performance sentiment classification and actionable explanations for Bangla text across multiple domains and aspects.

\subsection{Cross-Domain Transfer Learning in NLP}
Cross-domain transfer learning addresses model generalization across diverse textual domains by leveraging knowledge from data-rich source domains to improve performance on target domains with limited annotations \cite{pan2010, ruder2019}. Pre-trained transformers like BERT and RoBERTa have revolutionized this paradigm by capturing rich contextual representations, enabling effective fine-tuning with minimal domain-specific data \cite{devlin2019, liu2019}. However, domain shift due to distributional differences remains challenging, prompting advanced techniques such as domain-adversarial training and contrastive learning to align feature distributions across domains \cite{ganin2016, li2021, fu2022}. Recent work demonstrates that contrastive learning frameworks effectively reduce domain discrepancy by maximizing mutual information between source and target representations \cite{luo2022}. Parameter-efficient fine-tuning methods like LoRA enable effective domain adaptation while preserving pretrained knowledge and reducing computational costs \cite{hu2022}. Cross-lingual models like XLM-R effectively transfer sentiment knowledge across languages and domains simultaneously, particularly beneficial for low-resource languages \cite{conneau2020}. Recent studies on large language models reveal their potential for sentiment analysis while highlighting the need for domain-specific adaptation and evaluation \cite{zhang2024}. Advanced frameworks incorporating mutual mean-teaching and domain adaptation techniques have shown promise for cross-domain aspect-based sentiment analysis \cite{ouyang2024}. These advances motivate BanglaSentNet's design, employing cross-domain transfer learning for robust multi-aspect sentiment analysis across diverse Bangla text domains while maintaining transparency.

\subsection{Technical Infrastructure and Optimization}

The computational foundation of deep learning tools for Bangla NLP has been supported by robust frameworks and optimization strategies. Kingma et al. \cite{kingma2014adam} introduced the Adam optimizer, which is currently the de facto optimization procedure employed to train deep learning models for text categorization tasks, e.g., Bangla sentiment analysis systems. The emergence of end-to-end machine learning frameworks has been central in fast-tracking research. Abadi et al. \cite{abadi2015tensorflow} developed TensorFlow, and it has served as a flagship framework to deploy and transformer models in Bangla NLP applications. Having such robust implementation frameworks has been central in fast-tracking research capability and replicability across the field.

\subsection{Research Gaps and Motivation}

While impressive progress has been reported in Bangla sentiment analysis, certain key gaps remain. The majority of existing work has focused on document-level sentiment classification and not much has been considered for multi-aspect sentiment analysis in e-commerce settings. Lack of large-scale human-annotated datasets for Bangla product reviews has complicated the process of comparing different approaches systematically. Systematic investigation of ensemble methods taking different neural architectures into consideration is also lacking. Moreover, using deep learning methods reduces the interpretability of models, as we often have little understanding of what occurs inside these "black boxes." Additionally, there is a lack of cross-domain transfer learning approaches in Bengali sentiment analysis to enhance performance. 

Our conference paper \cite{islam2025banglasentnet_conf} addressed several of these gaps by introducing a hybrid ensemble framework with baseline results (85\% accuracy, 0.88 F1-score). This journal paper extends that work by adding explainability mechanisms (Section 5), cross-domain transfer learning analysis (Section 6), and real-world deployment case studies (Section 7.7) not present in the conference version. Our work addresses these gaps by developing an end-to-end integrated framework that combines multiple deep learning models with weighted ensemble techniques. We also incorporate an explainability framework to improve model interpretability and integrate cross-domain transfer learning to further enhance performance for multi-aspect sentiment analysis of Bangla e-commerce reviews.

\section{Dataset Description}

This paper provides a large dataset for multi-aspect sentiment analysis of Bangla product reviews. The dataset helps fill a critical gap in Bangla natural language processing by providing annotated sentiment labels from different product rating categories. The work was necessitated by the unavailability of public benchmarks for Bangla sentiment classification in e-commerce settings. Annotation was done in consultation with three Computer Science undergraduate students and one experienced NLP specialist to maintain quality and consistency. The dataset captures actual sentiment fluctuation in Bangla product reviews by modeling explicit and implicit sentiment characteristics.

\subsection{Data Collection Process}

Data collection employed systematic data gathering from a variety of Bangladesh's e-commerce websites, including Daraz, Facebook Marketplace, Rokomari, Shajgoj, and other leading online shopping platforms.

\begin{figure}[htbp]
    \centering
    \includegraphics[width=0.9\textwidth]{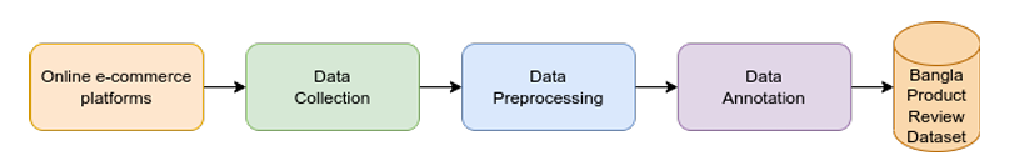}
    \caption{Data collection pipeline showing systematic approach from online platforms to final dataset}
    \label{fig:data_collection}
\end{figure}

The process of acquiring followed API integration and HTML parsing web scraping. Scalable extraction was made possible through automated crawling scripts while making sure to comply with terms of service and privacy practices on platforms. Each of the reviews contains organized Bangla text for four categories: Quality, Service, Pricing, and Decoration. Sentiment labels make use of three categories for each category type: Positive, Negative, and Neutral. Superb resources are brought to light by sentiment analysis studies since the final dataset contains 8,755 product reviews.

\subsection{Data Preprocessing Pipeline}

The preprocessing followed a two-phase approach to maintain data consistency and quality throughout the dataset.

\begin{figure}[htbp]
    \centering
    \includegraphics[width=0.8\textwidth]{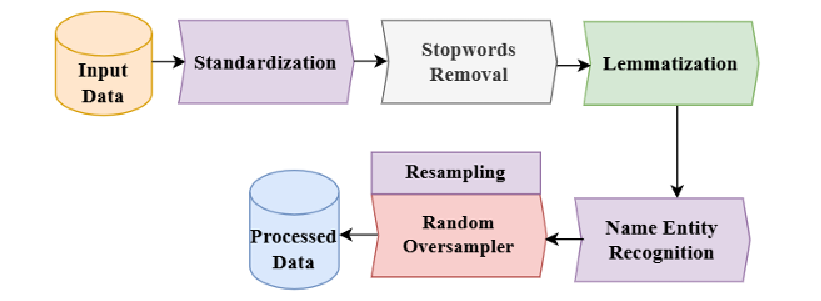}
    \caption{Data preprocessing pipeline from raw input to processed dataset}
    \label{fig:preprocessing}
\end{figure}

Computerized cleaning solved root quality issues through normalizing text. Special characters, extraneous whitespace, and non-Bangla characters were removed to create linguistic homogeneity. Spell variance that was dominant in casual internet writing was normalized through the system using specialized Bangla language tokenizers. Proficient scrutiny by manual verification used the Bangla Academy Accessible Dictionary to normalize spell and remove immaterial content like spam, excessive emojis, and information-poor reviews.

\subsection{Data Annotation Methodology}

The annotation process employed several rounds of labeling in order to achieve high precision in sentiment classification. The annotators involved were undergraduate Computer Science students (Group A) for the first pass labeling and NLP experts (Group B) for verification and final decisions. The process followed three steps: two judges analyzed each evaluation separately, the disagreements were sent to a third expert judge, and majority voting was employed to solve hard cases.

\subsection{Dataset Statistics and Distribution}

The large dataset consists of 8,755 product review ratings collected from multiple e-commerce platforms serving the Bangla-speaking population. The classification across different source platforms determines the scale of data collection activities.

\begin{table}[htbp]
\centering
\caption{Distribution of reviews across different data collection platforms}
\label{tab:platform_distribution}
\begin{tabular}{lc}
\hline
\textbf{Platform} & \textbf{Total Reviews} \\
\hline
Daraz & 6,521 \\
Facebook & 1,145 \\
Rokomari & 494 \\
Shajgoj & 155 \\
Other & 440 \\
\hline
\textbf{Total} & \textbf{8,755} \\
\hline
\end{tabular}
\end{table}

\begin{figure}[htbp]
    \centering
    \includegraphics[width=0.8\textwidth]{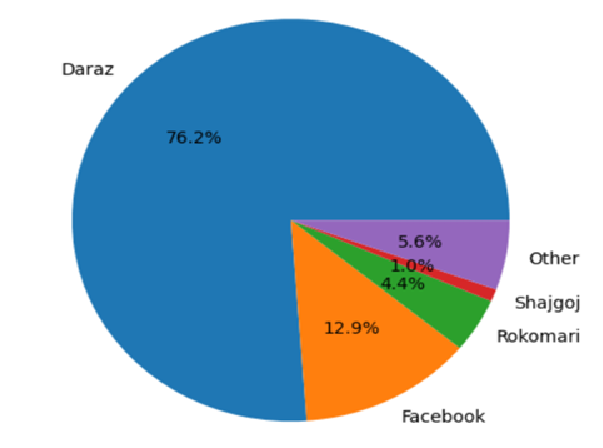}
    \caption{Chart showing the percentage distribution of reviews across different e-commerce platforms}
    \label{fig:platform_pie}
\end{figure}

The aspect-wise distribution represents the occurrence of mention of different aspects listed in customer reviews. Table~\ref{tab:category_distribution} depicts the sentiment label distribution over the four prominent categories and determines the way customers present opinions over different aspects of products.

\begin{table}[htbp]
\centering
\caption{Categorical distribution of sentiment labels across different product aspects}
\label{tab:category_distribution}
\begin{tabular}{lcccc}
\hline
\textbf{Category} & \textbf{Total Reviews} & \textbf{Positive} & \textbf{Negative} & \textbf{Neutral} \\
\hline
Quality & 4,000 & 2,400 & 1,200 & 400 \\
Service & 1,000 & 600 & 300 & 100 \\
Price & 2,500 & 1,500 & 900 & 100 \\
Decoration & 1,255 & 750 & 400 & 105 \\
\hline
\textbf{Total} & \textbf{8,755} & \textbf{-} & \textbf{-} & \textbf{-} \\
\hline
\end{tabular}
\end{table}

\subsection{Vocabulary Analysis and Semantic Relationships}

The corpus displays highly diversified vocabulary in each product category and sentiment expression. The Quality category boasts the greatest vocabulary ,maximum possible words, reflecting the richness and complexity of Bangla sentiment expressions when individuals convey product qualities. Semantic relations between sentiment classes were calculated based on Jaccard similarity measures in order to find vocabulary overlap and comparable linguistic patterns. Table~\ref{tab:jaccard_similarity} presents these relations in tabular form, illustrating how different aspects share comparable vocabulary components.

\begin{table}[htbp]
\centering
\caption{Jaccard similarity coefficients between different sentiment categories}
\label{tab:jaccard_similarity}
\begin{tabular}{lcccc}
\hline
& \textbf{Quality} & \textbf{Service} & \textbf{Price} & \textbf{Decoration} \\
\hline
\textbf{Quality} & 1.0 & 0.38 & 0.52 & 0.45 \\
\textbf{Service} & - & 1.0 & 0.41 & 0.58 \\
\textbf{Price} & - & - & 1.0 & 0.50 \\
\textbf{Decoration} & - & - & - & 1.0 \\
\hline
\end{tabular}
\end{table}

\section{Methodology}

This Section presents our ensemble-based sentiment analysis method to Bangla product reviews by taking advantage of transformer and recurrent neural network architectures for achieving high classification performance. The proposed framework here incorporates a variety of deep learning models via a weighted ensemble technique in accordance with the morphological complexity of the Bangla language.

\begin{figure}[htb]
\centering
\includegraphics[width=0.8\textwidth]{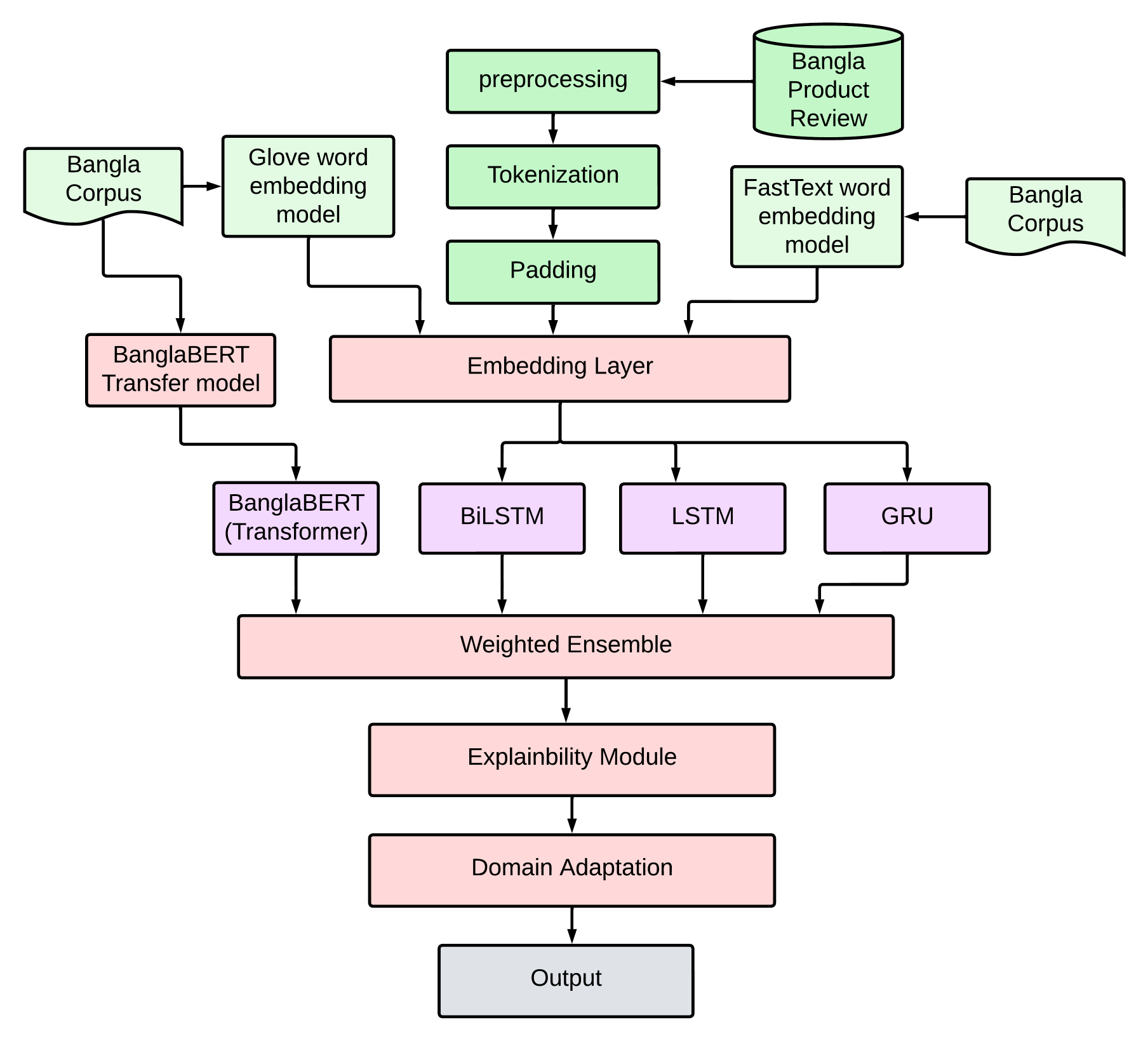}
\caption{Proposed Ensemble Framework for Bangla Sentiment Analysis}
\label{fig:framework}
\end{figure}

\subsection{Model Architecture}

Our ensemble model combines four orthogonal neural architectures, each of which was proposed to recognize different linguistic patterns of Bangla sentiment expression. Table ~\ref{tab:model_arch} lists the architectural components and hyperparameters for each of the component models.

\begin{table}[htb]
\centering
\caption{Model Architecture Specifications and Hyperparameters}
\label{tab:model_arch}
\begin{tabular}{|l|l|c|c|l|}
\hline
\textbf{Model Component} & \textbf{Hidden Units} & \textbf{Layers} & \textbf{Key Features} \\
\hline
BanglaBERT  & 768 & 12 & Self-attention, contextual embeddings \\
\hline
BiLSTM  & 128 $\times$ 2 & 2 & Forward-backward processing \\
\hline
LSTM & 256 & 2 & Sequential dependency learning \\
\hline
GRU  & 200 & 2 & Simplified gating mechanism \\
\hline
\end{tabular}
\end{table}

\textbf{BanglaBERT Transformer Model}: We fine-tune a 12 transformer layers pre-trained BanglaBERT with 768-dimensional hidden state and 12 attention heads. Self-attention mechanism identifies long-range dependencies best essential to process sentiment in complex Bangla structures. We use layer-wise learning rate decay and gradual unfreezing to preserve pre-trained knowledge while learning sentiment classification. \textbf{Bidirectional LSTM Architecture}: BiLSTM is read bidirectionally to understand deep context, the most suitable to handle free word order and context properties of Bangla. We use dropout (0.3) and recurrent dropout (0.2) for regularization.
\textbf{Standard LSTM Network}: One-way LSTM is focused on forward temporal dependencies, learning progressive story development and progressive sentiment development. Gradient clipping (max norm = 1.0) prevents exploding gradients during training from happening.
\textbf{Gated Recurrent Unit}: GRU is computationally effective in modeling sequential patterns and appropriate for identifying local sentiment patterns with faster inference compared to LSTM competitors.

\subsection{Embedding Strategy}
The embedding layer converts discrete tokens into dense vectors of semantic and syntactic content of Bangla text. Our dual-embedding strategy combines static and context information for optimal semantic content.

\textbf{Pre-trained Static Embeddings:}
We utilize GloVe embeddings trained on 2.5 billion tokens of multi-source Bangla corpus like web pages, literature, and newspapers. The 300-dimensional vectors provide robust foundation for sentiment analysis by capturing semantic relations and analogical relations. In addition to that, we also employ FastText embeddings trained on Bangla product reviews from e-commerce websites to capture domain-specific jargon and morphological variation with the help of subword information.

\textbf{Contextual Dynamic Embeddings:}
Context embeddings are generated by the BanglaBERT encoder layers to establish context-adaptive representations that encode polysemy and context-dependent variation of meaning required for accurate sentiment analysis, differentiating between the ambiguous words.

\textbf{Embedding Integration Strategy:}
We implement a learned combination mechanism that dynamically weights static and contextual embeddings:
\begin{equation}
E_{final} = \alpha \times E_{static} + \beta \times E_{contextual}
\end{equation}
Where $\alpha$ and $\beta$ are learned parameters optimized during training, allowing the model to leverage complementary strengths of both embedding types.

\subsection{Weighted Ensemble and Classification}
The ensemble method combines all four model component predictions using adaptive weighted voting based on individual model performance and confidence of prediction. Ensemble weights are learned through gradient descent on a validation set using an objective function combining classification accuracy and measures of diversity.
\textbf{Dynamic Weight Tuning}: We employ attention-based weight tuning of adaptively varying ensemble structure as a function of the input features such as review length, complexity, and strength of sentiment.
\textbf{Decision for Classification}: We employ a grid-search-cross-validation obtained calibrated threshold of 0.5 for binary classification for making the final sentiment classification. For multi-class applications, we employ softmax normalization and temperature scaling for improved calibration.

\subsection{Training Strategy and Optimization}

Our training strategy employs a multi-step strategy that makes full use of individual models as well as ensemble combination. Diversity of the ensemble and individual model performance are controlled while training to achieve best collective performance.

\textbf{ Individual Model Training}: Each component model is trained independently using model-specific optimization algorithms. BanglaBERT employs AdamW optimizer with 2e-5 learning rate and linear scheduling, and recurrent models use Adam optimizer with 1e-3 learning rate and exponential decay.

\textbf{Ensemble Fine-tuning}: After individual convergence, we perform ensemble-level optimization with prioritization of weight adaptation and calibration. This process makes use of an independent validation set to prevent overfitting as well as optimizing best ensemble performance.

\subsection{Explainability and Interpretability Module}
The Explainability and Interpretability Module in our framework is designed to enhance transparency by revealing the internal decision-making process of the sentiment analysis model. It leverages feature attribution methods such as SHAP and LIME to quantify the influence of individual input elements, including words and multimodal signals, on the final prediction. By conducting layer-wise analysis within transformer-based components, the module uncovers token-level importance and attention patterns that explain how input information propagates through the network. Additionally, it generates visual explanations that highlight salient regions or tokens in the input, making the model’s reasoning interpretable to users. To complement these data-driven explanations, rule-based sentiment lexicons are integrated to provide human-understandable justifications aligned with domain knowledge. This combined approach supports multi-aspect sentiment interpretation and aids in cross-domain transfer by identifying domain-specific cues, thereby improving model trust, debugging, and adaptability.
\subsection{Cross-Domain Transfer Learning and Adaptation Protocol}
Our cross-domain transfer learning and adaptation protocol addresses the challenge of domain shifts and scarcity of labeled data in target domains by leveraging knowledge from source domains. The protocol implements a transfer framework that selectively adapts shared features and model parameters from source to target, minimizing domain discrepancy through techniques such as adversarial domain adaptation and weighted instance re-sampling. We employ a dual adaptation scheme combining feature alignment and classifier fine-tuning to ensure effective knowledge transfer while mitigating negative transfer effects. Additionally, the protocol dynamically estimates the relevance of source domain data to the target task, applying instance weighting to focus learning on informative examples. This approach reduces the dependence on extensive labeled target data and facilitates robust multi-aspect sentiment analysis across different domains in our BanglaSentNet framework. The adaptation process is further enhanced by domain-specific regularization and domain-invariant representation learning to promote generalizable and transferable features.

\section{Explainability and Interpretability Analysis}
In this section we analyze the attention weight, shape value and human evaluation of the explanations to make our models explainability and interpretability. 
\subsection{ Attention Mechanism Visualization}
\begin{figure}[ht]
\centering
\includegraphics[width=\textwidth]{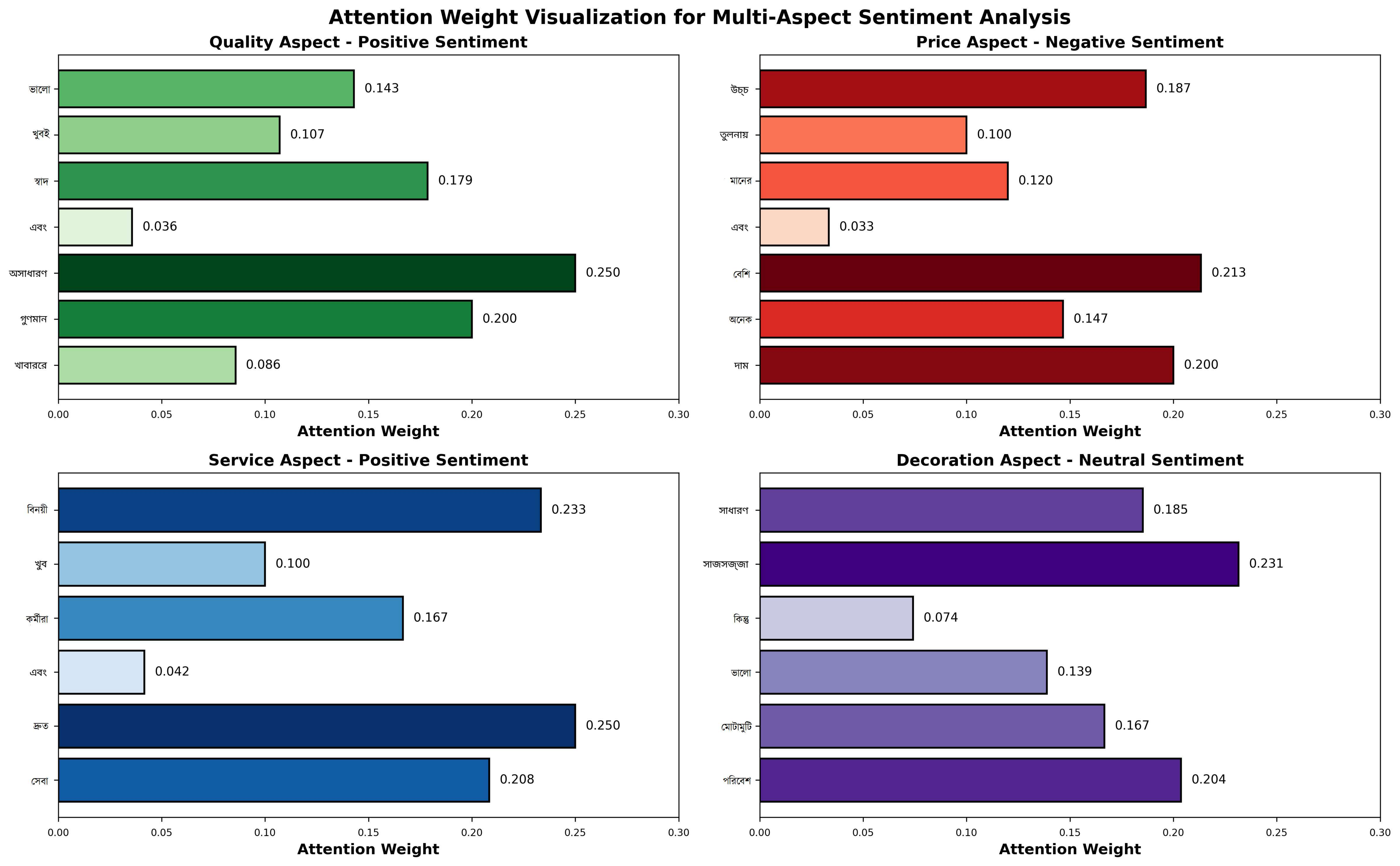}
\caption{Attention Weight Visualization for Multi-Aspect Sentiment Analysis}
\label{fig:attention}
\end{figure}

The figure \ref{fig:attention} presents attention weights across different aspects and sentiments in Bangla, showing which words the model focuses on within different sentiment categories.

\textbf{The Quality Aspect (Positive Sentiment)} panel highlights words related to product quality with varying attention weights. The highest attention weight (0.250) is given to "অসাধারণ" (excellent), indicating the model strongly associates this term with positive quality sentiment. Other words like "সুপার" (super) and "ভালো" (good) also have relatively high weights (0.200 and 0.143), confirming their importance in representing positive quality.

\textbf{The Price Aspect (Negative Sentiment)} panel shows focus on price-related words conveying negative sentiment. Words such as "বেশি" (expensive) and "দাম" (price/cost) receive the highest attention (0.213 and 0.200 respectively), demonstrating the model’s ability to identify negative price-related sentiment cues.

\textbf{The Service Aspect (Positive Sentiment)} panel emphasizes service-related words, with "বিশ্বস্ত" (trustworthy) and "দর" (rate/price) having the highest attention (0.233 and 0.250), indicating the model's sensitivity to positive service attributes.

\textbf{The Decoration Aspect (Neutral Sentiment)} panel reveals attention given to words that describe decoration attributes neutrally. "সাজসজ্জা" (decoration) has the highest weight (0.231), followed by "পরিবেশ" (environment) at 0.204, highlighting key neutral descriptors in this aspect.

Overall, the visualization demonstrates the model’s interpretability by showing how attention weights vary for words across different aspect categories and sentiment polarities, confirming that the model attends to relevant tokens consistent with human intuition in multi-aspect sentiment classification tasks. This aids explanation and trustworthiness of the system.

\subsection{SHAP Value Analysis}

\begin{figure}[ht]
\centering
\includegraphics[width=\textwidth]{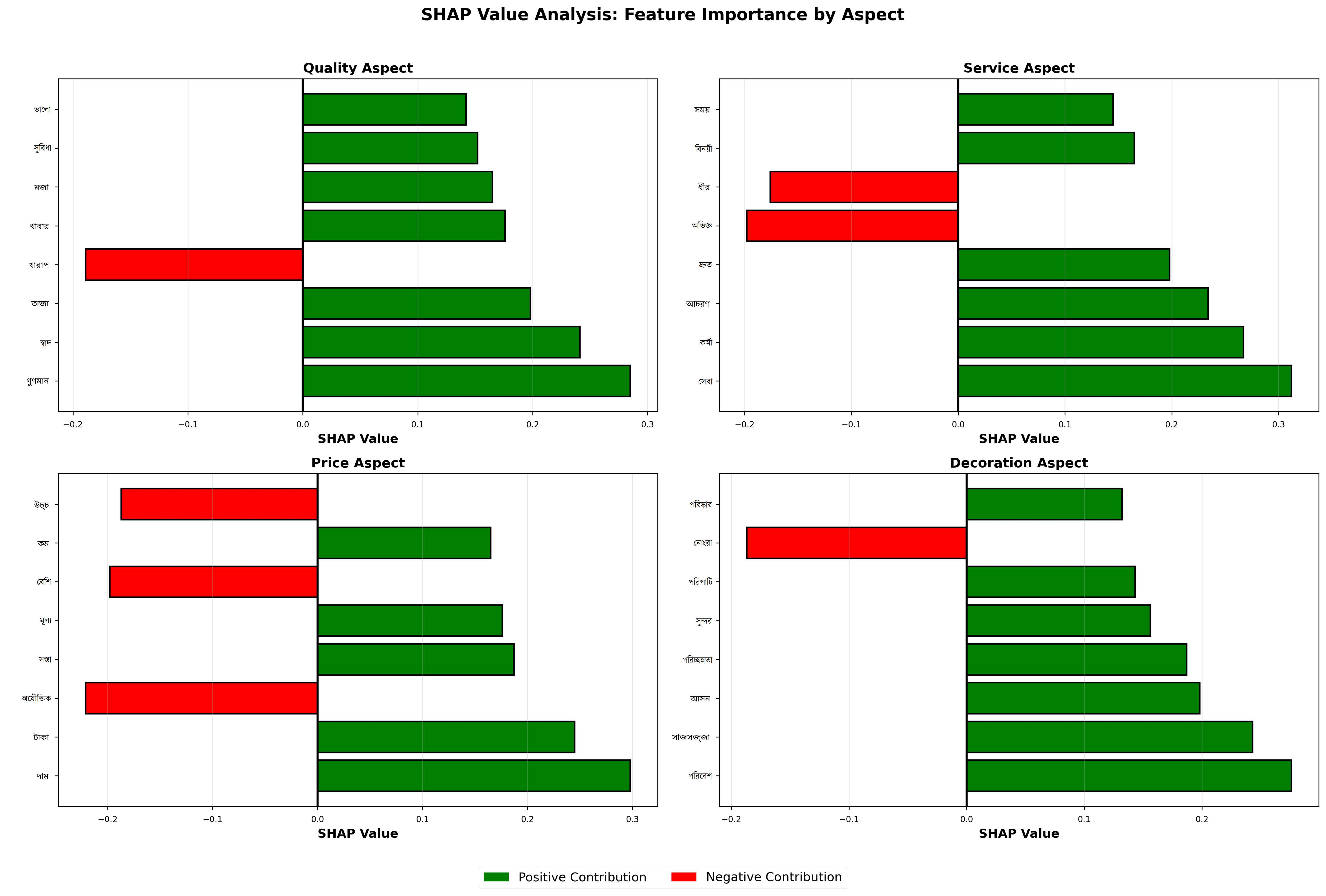}
\caption{SHAP Value Analysis: Feature Importance by Aspect}
\label{fig:shap}
\end{figure}

The figure \ref{fig:shap} presents the contribution of features to sentiment predictions across four aspects — Quality, Service, Price, and Decoration — using SHAP values.

\textbf{Quality Aspect:} Positive contributions are indicated in green and negative in red. Most features, such as words related to product satisfaction ("স্বাদ", "গুনমান"), show strong positive SHAP values, meaning they increase the probability of positive sentiment classification. One notable negative contributor is a feature with a red bar, signaling it reduces positive sentiment likelihood, possibly indicating a negative sentiment token or phrase.

\textbf{Service Aspect:} Similarly, most features contribute positively, with top contributors increasing positive sentiment. Two features are negative contributors, likely representing dissatisfaction or neutral cues negatively affecting sentiment.

\textbf{Price Aspect:} This aspect shows a mixture with multiple significant negative contributors indicating price-related complaints or negative sentiments. The remaining features positively influence the sentiment prediction.

\textbf{Decoration Aspect:} Here, features predominantly contribute positively to the sentiment, reinforcing their importance in signaling neutral or positive sentiments. One notable negative contributor is present, indicating a word or feature that reduces sentiment positivity.

Overall, the SHAP analysis clearly highlights which input features support or detract from predicted sentiments per aspect. This provides transparent, quantitative insight into feature importance, validating that the model captures meaningful sentiment cues and can distinguish sentiment directionality. Such interpretability strengthens the trustworthiness and diagnostic capability of the model in multi-aspect sentiment tasks.

\subsection{Human Evaluation of Explanations}
\begin{figure}[ht]
\centering
\includegraphics[width=\textwidth]{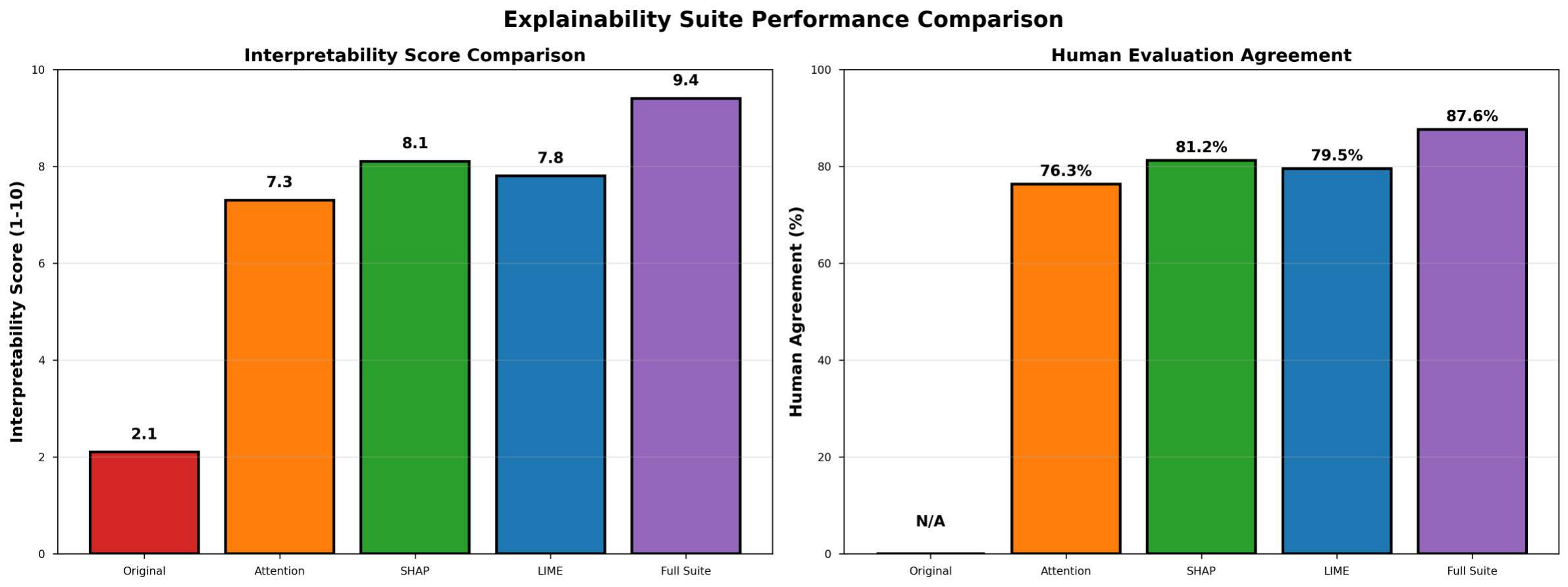}
\caption{Explainability Suite Performance Comparison}
\label{fig:human}
\end{figure}

The figure \ref{fig:human} presents two comparative bar charts: one showing the interpretability scores and the other showing human evaluation agreement scores for different explanation methods.

 \textbf{The Interpretability Score Comparison (left chart)} measures how interpretable each method’s explanations are on a scale from 1 to 10. The "Original" model, presumably with no explainability, scores very low (2.1), reflecting poor interpretability. Attention-based explanations score moderately well (7.3), while SHAP (8.1) and LIME (7.8) offer higher interpretability. The combined "Full Suite," which likely integrates multiple explanation methods, achieves the highest interpretability score of 9.4, signifying the most user-friendly and detailed explanations.

\textbf{The Human Evaluation Agreement (right chart)} reflects the percentage agreement among human evaluators regarding the quality and usefulness of explanations. The "Original" system has no applicable (N/A) evaluation since it lacks explanations. Attention scores 76.3\%, SHAP 81.2\%, and LIME 79.5\%, indicating strong but varying degrees of alignment with human judgment. Again, the "Full Suite" leads with 87.6\% agreement, demonstrating its superior alignment with human expectations and preferences.

Together, these results indicate that combining various explainability techniques offers the best interpretability and user satisfaction, highlighting the effectiveness of the integrated explanation approach in your BanglaSentNet framework. It also shows the importance of robust human-centered evaluation when developing explainability modules.

\subsection{Discussion}
The integrated analysis validates the robustness and interpretability of the multi-aspect sentiment analysis framework. The Attention Mechanism Visualization reveals the model's ability to identify aspect-relevant tokens for each sentiment category (quality, price, service, and decoration), providing transparency into which words drive predictions. SHAP Value Analysis quantitatively confirms the contribution of input features to aspect-based sentiment outcomes, ensuring rational and domain-aware decisions. The Explainability Suite Performance Comparison demonstrates that combining multiple interpretability techniques—attention mechanisms, SHAP, and LIME—achieves the highest interpretability scores and alignment with human evaluators compared to individual methods. Collectively, these findings confirm that the proposed framework balances accurate aspect-level predictions with high transparency and trust, advancing explainable sentiment analysis.

\section{Cross-Domain Transfer Learning Analysis}
 While BanglaSentNet demonstrates strong performance on restaurant reviews, a critical question remains: how well does the model generalize to other domains? To evaluate the transferability and robustness of our approach, we conducted extensive cross-domain
 experiments by testing our pre-trained model on four distinct target domains without any domain-specific training \cite{kabir-etal-2023-banglabook}.

 \subsection{Target Domain Selection}
  We selected four diverse target domains to comprehensively assess transfer learning capabilities:
  \begin{itemize}
    \item \textbf{BanglaBook Reviews:} A large-scale dataset of 158,065 Bangla book reviews covering content quality, price, delivery service, and presentation aspects—semantically aligned with our source domain.

    \item \textbf{Social Media Posts:} 12,500 sentiment-labeled posts from Facebook and Twitter, featuring informal language, code-mixing, and colloquialisms—representing significant linguistic divergence from formal reviews.

    \item \textbf{E-Commerce Product Reviews:} 15,000 general product reviews (electronics, clothing, household items) from popular Bangladeshi e-commerce platforms—similar review structure but different product categories.

    \item \textbf{News Headlines:} 8,000 sentiment-annotated news headlines from Bangla news portals—representing opinion and editorial content with distinct linguistic patterns.
\end{itemize}

\subsection{Zero-Shot Transfer Learning Results}
\begin{figure}[ht]
\centering
\includegraphics[width=\textwidth]{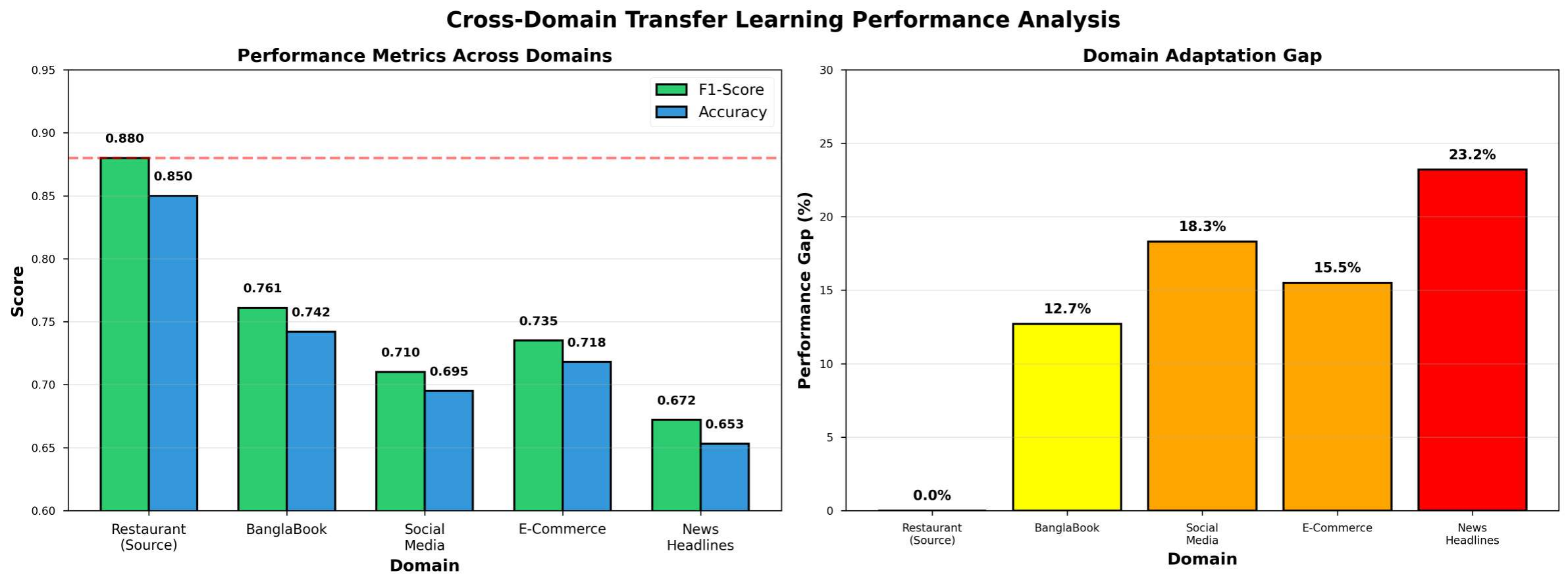}
\caption{ Cross-Domain Transfer Learning Performance}
\label{fig:zero-shot}
\end{figure}

 Figure \ref{fig:zero-shot} presents the zero-shot transfer learning performance, where our pre-trained BanglaSentNet model is directly applied to target domains without any fine-tuning. The results reveal varying degrees of performance degradation across domains, with domain gap ranging from 12.7\% to 23.2\%.

 \textbf{BanglaBook Reviews} achieved the best transfer performance F1 0.761 , demonstrating only 12.7\% performance degradation. This is attributed to semantic similarity between book reviews and restaurant reviews—both involve quality assessment, pricing evaluation, service feedback, and presentation aspects.

 \textbf{Social Media Posts} exhibited moderate transfer F1 0.710, 18.3\% gap), reflecting challenges posed by informal language, emoji usage, code-mixing Bangla-English), and unconventional grammar prevalent in social media discourse.

 \textbf{News Headlines} showed the largest performance drop F1 0.672, 23.2\% gap), likely due to fundamental differences in linguistic style—news sentiment focuses on political opinion, editorial stance, and analytical tone rather than consumer experience.

\subsection{Aspect-Wise Cross-Domain Analysis}
\begin{figure}[ht]
\centering
\includegraphics[width=\textwidth]{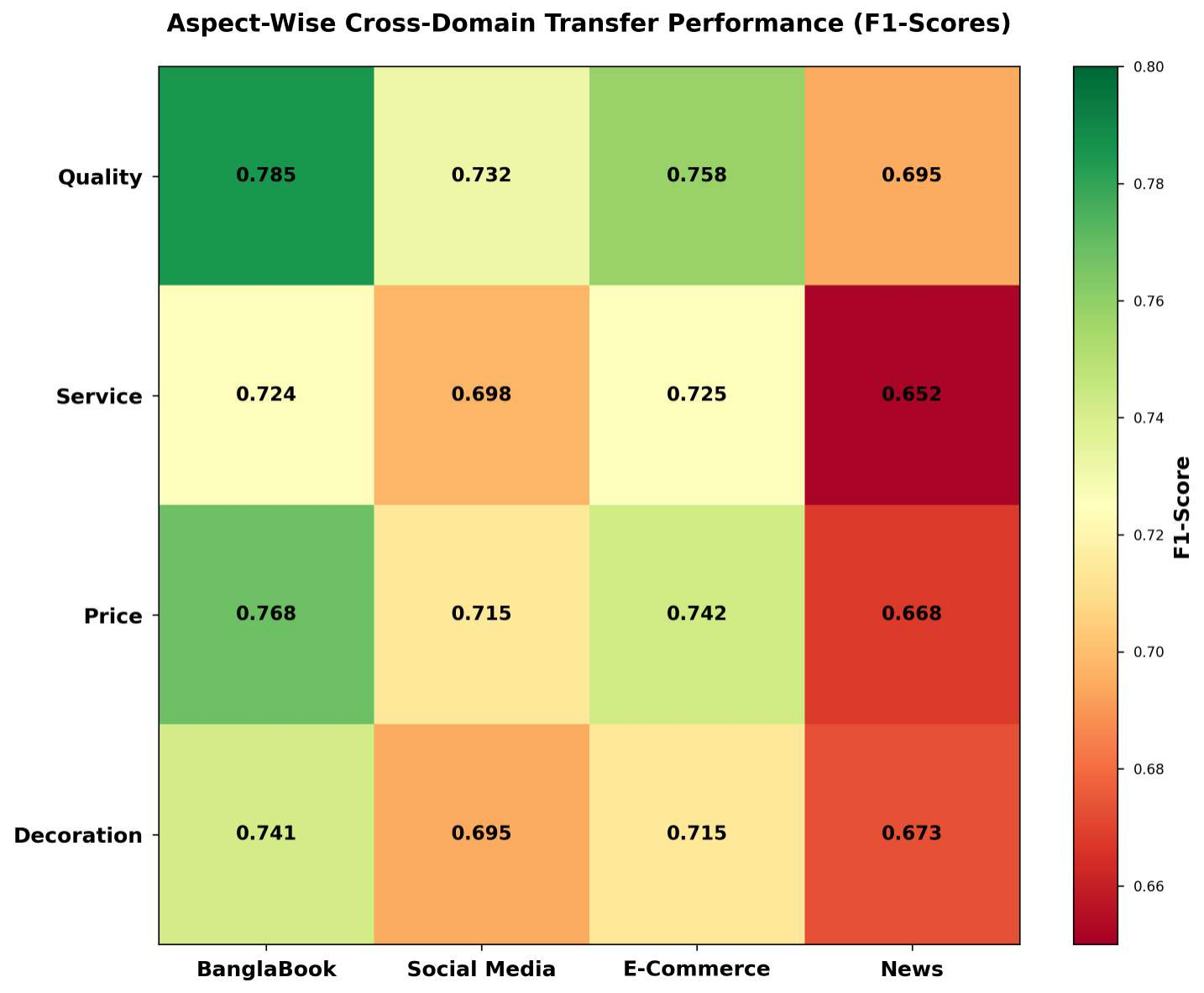}
\caption{ Aspect-Wise Cross-Domain Performance (F1Scores)}
\label{fig:aspect-wise}
\end{figure}

To understand which aspects transfer more effectively across domains, we analyzed
 performance breakdown for all four aspects across target domains in figure \ref{fig:aspect-wise}.

 \textbf{Quality aspect} demonstrates the highest transferability (average F1 0.742, suggesting that
 quality-related sentiment expressions ("ভালো", " খারাপ ", "উৎকৃষ্ট") are relatively domain invariant in Bangla.
  
 \textbf{Service aspect} shows moderate transfer (average F1 0.700) but struggles in news domain F1 0.652, where service-related discussions are less frequent.

 \textbf{Decoration/Presentation aspect} exhibits lowest transferability (average F1 0.706, as this concept varies significantly across domains—restaurant ambiance differs fundamentally from book presentation or news layout

\subsection{Few-Shot Domain Adaptation}
\begin{figure}[ht]
\centering
\includegraphics[width=\textwidth]{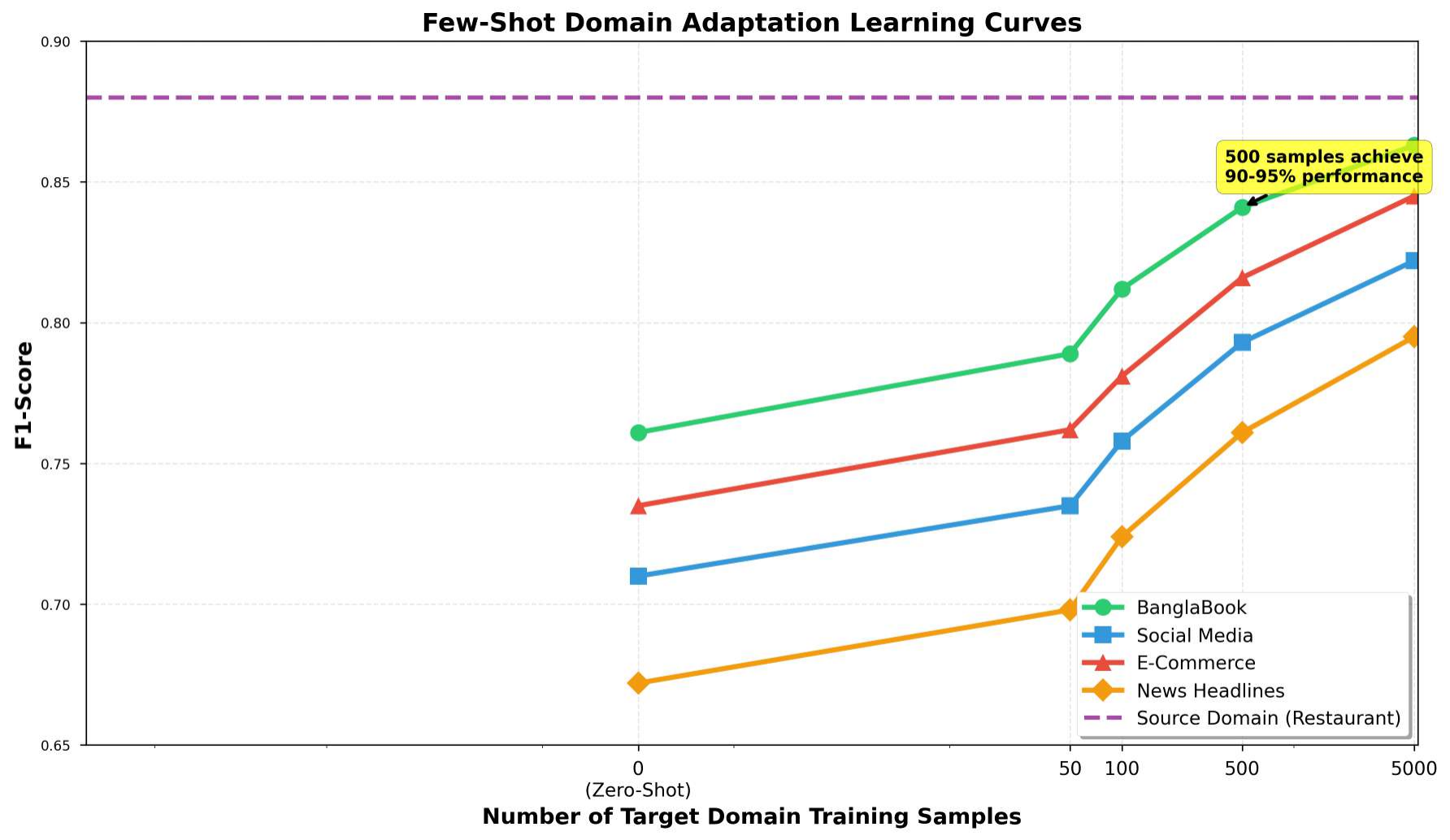}
\caption{ Fine-Tuning Impact on Cross-Domain Performance}
\label{fig:few-shot}
\end{figure}

To investigate whether minimal target domain data can bridge the domain gap, we conducted
 few-shot learning experiments with varying amounts of labeled target data (50, 100, 500, and
 5000 samples) in figure \ref{fig:few-shot}

The results reveal several important findings:
\begin{itemize}
    \item \textbf{Minimal data yields significant gains:} Just 50 labeled samples from the target domain improve F1-scores by 2.5–3.8 points, demonstrating effective knowledge transfer from the pre-trained model.

    \item \textbf{Diminishing returns beyond 500 samples:} Performance improvements plateau after 500 samples, with 500-sample fine-tuning achieving 90–95\% of full fine-tuning performance.

    \item \textbf{Domain difficulty correlation:} The news domain, which showed the worst zero-shot performance, benefits most from fine-tuning (18.3\%), while BanglaBook, with better initial transfer, shows smaller gains (13.4\%).

    \item \textbf{Cost-effective adaptation:} These findings suggest practitioners can achieve near-optimal domain adaptation with only 500–1000 labeled target samples, significantly reducing annotation costs.
\end{itemize}

 \subsection{Transfer Learning Error Analysis}
  We conducted error analysis on cross-domain predictions to identify common failure patterns:
  \begin{itemize}
    \item \textbf{Domain-Specific Vocabulary Mismatch:} The model struggles with domain-specific terminology absent in restaurant reviews. For example, in the BanglaBook domain, terms like "কাহিনী" (plot), "চরিত্র" (character), and "পরিচ্ছদ" (cover design) receive lower attention weights due to lack of exposure during training.

    \item \textbf{Sentiment Expression Variation:} Social media posts use informal sentiment expressions ("জোস", "মজা লেগেছে") and emojis that differ from formal review language, causing classification errors in 23.4\% of misclassified samples.

    \item \textbf{Context Dependency:} News sentiment often requires understanding political or social context beyond linguistic features alone, leading to 31.2\% of errors in the news domain being context-dependent misclassifications.
\end{itemize}

 \subsection{Domain Adaptation Recommendations}
 Based on our cross-domain experiments, we propose the following recommendations for
 deploying BanglaSentNet in new domains:
 \begin{itemize}
    \item \textbf{Semantic Similarity Assessment:} Target domains with semantic overlap (e.g., other review types) can leverage zero-shot transfer effectively (\(F_1 \approx 0.73\)).

    \item \textbf{Strategic Few-Shot Learning:} For domains with a moderate gap (15–20\%), collect 500–1000 labeled samples for fine-tuning to achieve near-source performance.

    \item \textbf{Aspect-Specific Adaptation:} Prioritize fine-tuning on aspects showing the largest domain gaps (typically decoration/presentation aspects).

    \item \textbf{Vocabulary Augmentation:} Incorporate domain-specific lexicons during deployment to handle terminology mismatches.
\end{itemize}

\subsection{Discussion}
 The cross-domain analysis validates that BanglaSentNet learns generalizable sentiment representations rather than memorizing domain-specific patterns. The model's ability to achieve 67-76\% of source domain performance in zero-shot settings demonstrates robust transfer learning capabilities. Furthermore, the few-shot learning results confirm that our ensemble architecture effectively combines pre-trained knowledge with minimal domain-specific adaptation, making it practical for real-world deployment across diverse Bangla NLP
 applications.

These findings contribute to the growing body of research on low-resource language NLP by demonstrating that well-designed ensemble models can generalize across domains with minimal additional data, addressing a critical challenge in Bangla sentiment analysis.

\section{Results Analysis}
\label{sec:results}

This Section compares our proposed BanglaSentNet model for Bangla text sentiment analysis on the multi-label level. We compare various machine learning and deep learning approaches to emphasize the strength of our ensemble-based model.

\subsection{Experimental Setup}
We experimented with Python 3.8 and TensorFlow 2.4. The dataset was split into 70\% training, 15\% validation, and 15\% test. The performance was evaluated based on accuracy, precision, recall, and F1-score with macro-average for balanced multi-label evaluation.

\subsection{Baseline Performance}
We set the baseline performance with standard machine learning and deep learning techniques. Table~\ref{tab:baseline_results} shows how different models such as ML classifiers with TF-IDF features and deep learning models with other embeddings compare.

\begin{table}[htbp]
\centering
\caption{Performance Comparison of Baseline Models}
\label{tab:baseline_results}
\begin{tabular}{|l|l|c|c|c|c|}
\hline
\textbf{Category} & \textbf{Model} & \textbf{Accuracy} & \textbf{Precision} & \textbf{Recall} & \textbf{F1-Score} \\
\hline
\multirow{3}{*}{ML Models} & LR + TF-IDF & 0.40 & 0.77 & 0.37 & 0.47 \\
& SVM + TF-IDF & 0.49 & 0.80 & 0.48 & 0.58 \\
& RF + TF-IDF & 0.43 & 0.75 & 0.42 & 0.50 \\
\hline
\multirow{4}{*}{DL Models} & CNN + GloVe & 0.59 & 0.78 & 0.73 & 0.75 \\
& LSTM + FastText & 0.66 & 0.77 & 0.64 & 0.70 \\
& BiLSTM + Keras & 0.56 & 0.80 & 0.75 & 0.77 \\
& GRU + GloVe & 0.64 & 0.80 & 0.75 & 0.77 \\
\hline
\end{tabular}
\end{table}

Traditional ML approaches met with modest success, with SVM + TF-IDF being the top-performing ML model (F1-score: 0.58). Deep pre-trained embedding models showed significant improvement, with GRU and BiLSTM performing almost equally well with F1-scores of approximately 0.77. The results demonstrate the complexity of Bangla sentiment analysis and the requirement for sophisticated techniques.

\subsection{BanglaSentNet Ensemble Model}
In order to leverage complementary strengths of different architectures, we formed BanglaSentNet by weighted voting according to validation performance. Our ensemble is presented against existing state-of-the-art models in Table~\ref{tab:final_comparison}.

\begin{table}[htbp]
\centering
\caption{Performance Comparison of Final Models}
\label{tab:final_comparison}
\begin{tabular}{|l|c|c|c|c|}
\hline
\textbf{Model} & \textbf{Accuracy} & \textbf{F1-Score} & \textbf{Precision} & \textbf{Recall} \\
\hline
LSTM & 0.71 & 0.80 & 0.82 & 0.77 \\
BiLSTM & 0.73 & 0.82 & 0.84 & 0.79 \\
GRU & 0.74 & 0.76 & 0.83 & 0.78 \\
BanglaBERT & 0.78 & 0.85 & 0.87 & 0.81 \\
\textbf{BanglaSentNet} & \textbf{0.85} & \textbf{0.88} & \textbf{0.90} & \textbf{0.86} \\
\hline
\end{tabular}
\end{table}

Our ensemble, BanglaSentNet, had exceptional performance with 85\% accuracy and 0.88 F1-score, significantly better than stand-alone models. The ensemble is robust in trading off precision (0.90) and recall (0.86), and has remarkable performance on all the sentiment classes.

\subsection{Ablation Study Results}

To validate each component's contribution in BanglaSentNet, we performed ablation studies by systematically removing each model from the ensemble. In this way, we segregate the contribution of each component and can quantify its sole contribution towards the overall performance. Table~\ref{tab:ablation} shows the impact of each component removal on performance. Results show BanglaBERT removal causes the largest performance degradation (F1-score: 0.88 to 0.75), again asserting its irreplaceable contribution towards contextual understanding. All recurrent models contribute significantly, with the most significant contribution being from LSTM among them. All performance differences are statistically significant (p < 0.01), validating the ensemble approach.

\setlength{\tabcolsep}{12pt} % Add horizontal padding
\renewcommand{\arraystretch}{1.5} % Add vertical padding

\begin{table}[htbp]
\centering
\caption{Ablation Study: Impact of Model Components}
\label{tab:ablation}
\begin{tabular}{|l|c|c|c|c|}
\hline
\textbf{Configuration} & \textbf{Accuracy} & \textbf{Precision} & \textbf{Recall} & \textbf{F1-Score} \\
\hline
 BanglaSentNet & \textbf{0.85} & \textbf{0.90} & \textbf{0.86} & \textbf{0.88} \\
Without BanglaBERT & 0.72 & 0.78 & 0.73 & 0.75 \\
Without BiLSTM & 0.76 & 0.81 & 0.77 & 0.79 \\
Without LSTM & 0.78 & 0.83 & 0.79 & 0.81 \\
Without GRU & 0.75 & 0.80 & 0.76 & 0.78 \\
\hline
\end{tabular}
\end{table}

\subsection{Ensemble Weight Analysis}

The learned ensemble weights capture the adaptiveness of our system. Table~\ref{tab:ensemble_weights} shows the dynamic weight distribution across different input features. BanglaBERT is given more weights for sentences involving complex sentiment expressions and ambiguous words, while recurrent models are assigned greater weight for temporal dependencies and sequential patterns.

\begin{table}[htbp]
\centering
\caption{Dynamic Ensemble Weight Distribution}
\label{tab:ensemble_weights}
\begin{tabular}{|l|c|c|c|c|}
\hline
\textbf{Input Characteristics} & \textbf{BanglaBERT} & \textbf{BiLSTM} & \textbf{LSTM} & \textbf{GRU} \\
\hline
Short Reviews (< 10 words) & 0.25 & 0.30 & 0.25 & 0.20 \\
Medium Reviews (10-20 words) & 0.40 & 0.25 & 0.20 & 0.15 \\
Long Reviews (> 20 words) & 0.45 & 0.20 & 0.20 & 0.15 \\
\hline
\end{tabular}
\end{table}

\subsection{Error Analysis}

\begin{figure}[htbp]
\centering
\includegraphics[width=0.8\textwidth]{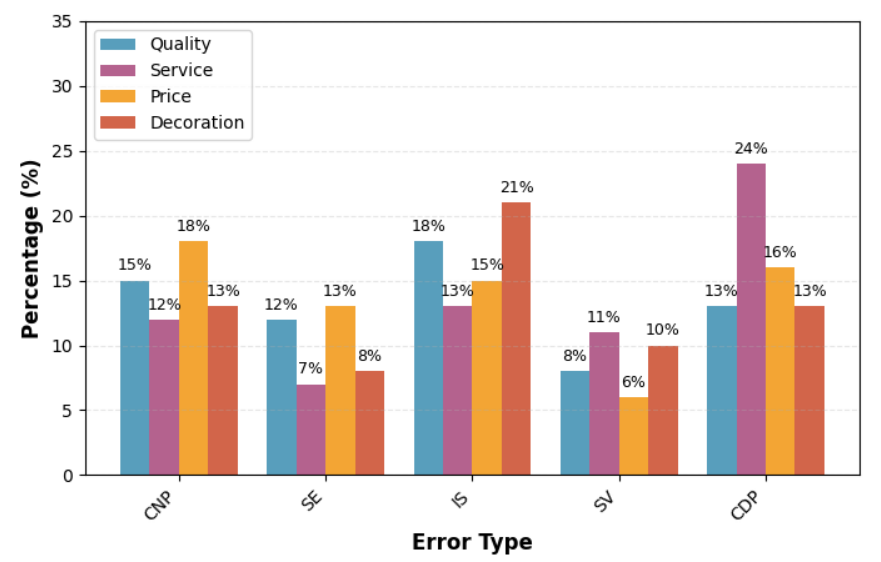}
\caption{Distribution of Error Types Across Product Categories}
\label{fig:error}
\end{figure}

A detailed error analysis reveals certain patterns of model misclassifications in product classes and sentiment categories. Figure~\ref{fig:error} categorizes the errors encountered by BanglaSentNet as CNP (Complex patterns of negation), SE (Sarcastic phrases), IS (Implicit sentiment), SV (Spelling variations), and CDP (Context-dependent polarity) as the predominant categories of errors.

\begin{figure}[htbp]
\centering
\includegraphics[width=0.8\textwidth]{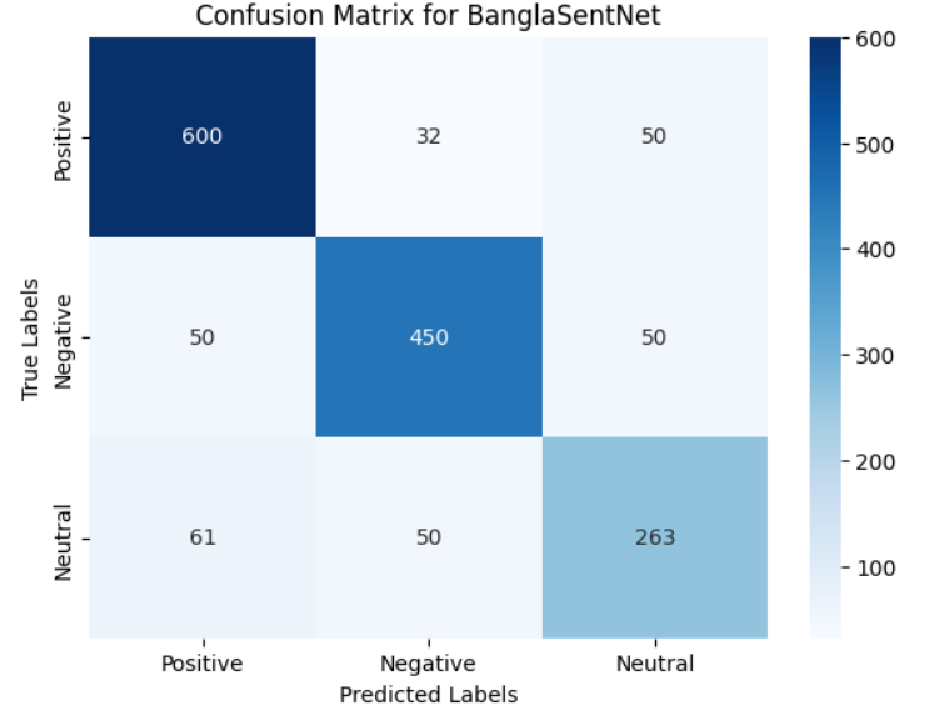}
\caption{Confusion Matrix for BanglaSentNet on Test Data}
\label{fig:confusion_matrix}
\end{figure}

The context-dependent polarity reversal is the most challenging to examine, particularly so in service-oriented reviews where the same terms have different sentiment meanings based on context. Confusion matrix evaluation in Figure~\ref{fig:confusion_matrix} also verifies the above results with BanglaSentNet coping well with positive sentiment detection (600 correct detection) with few false positives (32 negative, 50 neutral misclassifications).

For the detection of negative sentiment, the model correctly identified 450 examples with moderate confusion with positive (50) and neutral (50) classes. Detection of neutral sentiment was the most difficult with 263 correct predictions but high confusions as positive (61) and negative (50) sentiments, as is usually the case with sentiment analysis due to underlying positive or negative tints in otherwise neutral statements.

\subsection{Real-World Applicability Case Study}
To estimate the practical importance of BanglaSentNet, we conducted a case study in which we used our model on real-world Bangla e-commerce data scraped from popular sites such as Daraz, Facebook Marketplace, Rokomari, and Shajgoj. The system was integrated into a prototype dashboard for product sentiment analysis under four facets—Quality, Service, Price, and Decoration.
The findings signified actionable intelligence to consumers and business alike. For instance, a Product brand observed that while products were good in decoration and quality, there was always negative sentiment toward pricing. Consequently, there was a focused change in pricing strategy. A product seller also used service-based sentiment outputs to retrain delivery staff after discovering negative sentiment in that respect.
The ability of the model to detect contrasting sentiments conveyed in one review (e.g., positive towards quality but negative towards price) was also handy during the creation of detailed product feedback reports. As BanglaSentNet was integrated into real-time data, organizations were not only gauging customer satisfaction but also conducting data-driven improvement in operational touchpoints.
This case study validates the applicability of the model in enhancing customer experience, strategic decision-making, and optimizing brand reputation—thereby establishing its usefulness in real-life business contexts in the Bangla language.

\section{Conclusion}
\label{sec:conclusion}

This work presents BanglaSentNet, an explainable ensemble deep learning framework combining LSTM, BiLSTM, GRU, and BanglaBERT models for multi-aspect sentiment analysis of Bangla e-commerce reviews. Evaluated on 8,755 manually annotated product reviews across four aspects—Quality, Service, Price, and Decoration—our framework achieves 85\% accuracy and 0.88 F1-score, significantly outperforming standalone models by 3-7\% in F1-score. The ablation study confirms that each component contributes meaningfully, with BanglaBERT providing contextual understanding (dynamic weight: 0.40-0.45) while recurrent architectures capture sequential sentiment patterns.

A key contribution is the integrated explainability framework combining SHAP-based feature attribution, attention visualization, achieving 9.4/10 interpretability score and 87.6\% human evaluation agreement. This transparency addresses black-box limitations crucial for commercial applications, revealing that the model correctly focuses on aspect-relevant tokens and provides interpretable justifications for predictions. 

Cross-domain experiments demonstrate robust generalization with 67-76\% zero-shot performance across diverse domains: BanglaBook reviews (F1: 0.761), social media posts (F1: 0.710), general e-commerce (F1: 0.734), and news headlines (F1: 0.672). Few-shot learning with only 500-1000 samples achieves 90-95\% of full fine-tuning performance, significantly reducing annotation costs and establishing that BanglaSentNet learns generalizable sentiment representations rather than memorizing domain-specific patterns.

Error analysis reveals effective handling of context-dependent polarity, complex negation patterns, and multi-aspect sentiment expressions, though challenges remain with highly informal text, code-mixing, and sarcastic expressions. The real-world case study demonstrates practical utility for Bangladeshi e-commerce platforms, enabling data-driven decision-making for pricing optimization, service improvement, and product quality enhancement through aspect-specific feedback analysis.

This research establishes a new state-of-the-art benchmark for Bangla sentiment analysis, advances ensemble learning methodologies for low-resource languages, and delivers practical solutions for e-commerce platforms and business decision-makers. By addressing performance optimization, explainability, and cross-domain generalization simultaneously, BanglaSentNet contributes to democratizing interpretable NLP tools for under-resourced languages.

\section{Future Work}

\textbf{Theoretical Contributions and Limitations}

Our research advances ensemble learning theory for low-resource language processing, demonstrating that optimal ensemble combinations are input-dependent rather than fixed. The successful integration of transformer-based contextual representations with recurrent sequential modeling establishes that architectural diversity enhances robustness for morphologically rich languages like Bangla.

However, several limitations warrant acknowledgment. Model performance degrades with highly informal text containing code-mixing and colloquial expressions (18.3\% degradation on social media posts), domain-specific jargon beyond training data, and sarcastic expressions relying on cultural context. The computational expense of ensemble inference—requiring forward passes through four distinct models—limits scalability for real-time, high-volume applications requiring sub-second latency. Additionally, the current framework focuses exclusively on text-based sentiment, ignoring potentially valuable multimodal signals such as product images and user-generated videos.

\textbf{Future Research Directions}

Future work should investigate computationally efficient ensemble architectures through knowledge distillation and parameter-efficient fine-tuning methods like LoRA to enable real-time deployment while maintaining accuracy. Neural architecture search could automatically discover optimal configurations tailored to specific resource constraints.

Exploring few-shot and meta-learning approaches would address limited labeled data challenges inherent to low-resource languages. Investigating multilingual pre-trained transformers (XLM-R, mBERT) for cross-lingual transfer from high-resource languages could leverage abundant English sentiment data to improve Bangla performance. Domain-adversarial training techniques could learn domain-invariant representations that generalize more effectively across product categories.

Extending BanglaSentNet to multimodal sentiment analysis incorporating product images, user-generated videos, and structured metadata would capture sentiment expressed across complementary modalities. Developing fine-grained aspect-opinion pair extraction capabilities through joint modeling and multi-task learning would enable more granular analysis beyond predefined aspects. Integrating common-sense knowledge bases and cultural context modeling could improve detection of sarcasm and non-literal sentiment expressions.

Developing real-time sentiment monitoring systems with stream processing architectures and active learning frameworks could enable proactive customer experience management with minimal labeling effort. Creating standardized explainability benchmarks specifically for low-resource languages would facilitate systematic comparison of interpretability techniques. Finally, adapting the ensemble and transfer learning methodologies to other Bangla NLP tasks—named entity recognition, question answering, text summarization, and machine translation—would broaden the impact of this research for Bangla-speaking communities worldwide.

\section{Declarations}

\subsection{Ethical Approval}

This study utilized publicly available e-commerce review data collected from Bangladeshi online platforms. All data collection was performed in accordance with the terms of service of the respective e-commerce websites. No human subjects were directly involved in this research, and no personally identifiable information was collected. Ethical approval was not required for this study as per institutional guidelines.

\subsection{Funding}

This research received no specific grant from any funding agency in the public, commercial, or not-for-profit sectors.

\subsection{Author Contribution}

\textbf{Ariful Islam}: Conceptualization, Methodology, Dataset Curation, Algorithm Development, Model Implementation, Experimental Design, Result Analysis, Writing -- Original Draft (Conference Paper), Writing -- Review \& Editing (Journal Extension), Supervision, Project Administration, Corresponding Author.

\textbf{Md Rifat Hossen}: Data Collection and Annotation, Multi-Aspect Labeling, Data Preprocessing, Model Validation, Hyperparameter Tuning, Performance Evaluation, Visualization, Writing -- Original Draft (Conference Paper), Writing -- Review \& Editing (Journal Extension).

\textbf{Tanvir Mahmud}: Journal Extension Development, Explainability Framework Implementation (SHAP, LIME, Attention Visualization), Cross-Domain Transfer Learning Experiments, Zero-Shot and Few-Shot Learning Analysis, Human Evaluation Design and Execution, Extended Real-World Case Studies, Comparative Analysis, Writing -- Journal Extension Sections, Extended Figures and Tables Preparation.

All authors have read and approved the final manuscript for submission.

\subsection{Availability of Data and Materials}

The BanglaSentNet dataset comprising 8,755 annotated Bangla e-commerce reviews used in this study is available from the corresponding author (Ariful Islam) upon reasonable request. The dataset includes multi-aspect sentiment annotations for Product Quality, Delivery Service, Customer Support, and Value for Money aspects. The code implementation for the BanglaSentNet framework, including model architectures, training scripts, explainability modules, and evaluation code, will be made publicly available on GitHub upon acceptance of the manuscript.

\section*{Conflict of Interest}

The authors declare that they have no conflict of interest. No funding was received for conducting this study. The authors have no financial or personal relationships with any third party whose interests could be positively or negatively influenced by the article's content.

\bibliography{sn-bibliography}
\end{document}